\documentclass[letterpaper]{article}
\usepackage{aaai}
\usepackage{times}
\usepackage{helvet}
\usepackage{courier}
\usepackage{url}
\usepackage{amsmath,amsthm, amssymb}
\usepackage{graphics}
\usepackage{amsfonts}
\usepackage{graphpap}
\usepackage{multicol}
\usepackage{graphicx}
\usepackage{extarrows}
\usepackage{mathtools}
\usepackage{rotating}
\usepackage{multicol}
\usepackage{color}
\usepackage{tikz}
\usepackage{relsize}
\usepackage{mdwlist}
\usepackage{enumerate}
\usepackage{cancel}
\usetikzlibrary{shapes,arrows,matrix}
\usepackage{stmaryrd}
\usepackage{listings}
\usepackage{hyperref}
\usepackage{tipa}
\usepackage{upgreek}
\usepackage{mathrsfs}
\usepackage{pdflscape}

\newtheorem{definition}{Definition}
\newtheorem{proposition}{Proposition}
\newtheorem{lemma}{Lemma}
\newtheorem{corollary}{Corollary}
\newtheorem{example}{Example}
\begin{document}
%
\title{Automatic Synthesis of Geometry Problems for an Intelligent Tutoring System}
\author{
  Chris Alvin \\
  Louisiana State University\\
  calvin1@lsu.edu 
  \And
  Sumit Gulwani \\
  Microsoft Research, Redmond\\
  sumitg@microsoft.com \\
  \And
  Rupak Majumdar \\
  MPI-SWS\\
  rupak@mpi-sws.org \\
  \And
  Supratik Mukhopadhyay \\
  Louisiana State University\\
  supratik@csc.lsu.edu \\
}


\maketitle

\begin{abstract}
This paper presents an intelligent tutoring system, \emph{GeoTutor}, for Euclidean Geometry that is automatically able to synthesize proof problems and their respective solutions given a geometric figure together with a set of properties true of it. GeoTutor can provide personalized practice problems that address student deficiencies in the subject matter.
\end{abstract}

\section{Introduction}

We present an intelligent tutoring system for Euclidean Geometry that can

\begin{itemize} 

\item Automatically synthesize personalized proof problems based on a student's background and performance history, 



\item Automatically provide practice problems that address student deficiencies in the subject matter. 

\end{itemize} 

Our system \emph{GeoTutor} takes as input a geometry problem given as a figure together with a set of properties true of the figure (e.g., two line segments are equal, or that an angle is 90 degrees). Internally, it represents the figure as a hypergraph whose nodes are basic propositions about geometric objects and whose hyperedges represent deductions that follow from the propositions using axioms of Euclidean geometry. We provide an algorithm to traverse the hypergraph of deductions to come up with problems related to the given template problem. Our traversal algorithm filters the state space using additional assumptions about the student's history or heuristics to limit to \emph{interesting} deductions. 

\begin{figure}
\begin{center}
\begin{tikzpicture}[line cap=round,line join=round,>=triangle 45,x=1.0cm,y=1.0cm,scale=0.65]
\definecolor{zzttqq}{rgb}{0.6,0.2,0}
\definecolor{qqqqff}{rgb}{0,0,1}
\definecolor{uuuuuu}{rgb}{0.27,0.27,0.27}
\definecolor{xdxdff}{rgb}{0.49,0.49,1}
\clip(-1,-0.1) rectangle (5,3.5);
\fill[color=zzttqq,fill=zzttqq,fill opacity=0.1] (3.6,0) -- (4,0) -- (4,0.4) -- (3.6,0.4) -- cycle;
\draw (0,3)-- (0,0);
\draw (0,3)-- (4,0);
\draw (4,3)-- (4,0);
\draw (4,3)-- (0,0);
\draw (0,0)-- (4,0);
\draw (3.6,0)-- (4,0);
\draw (4,0)-- (4,0.4);
\draw (4,0.4)-- (3.6,0.4);
\draw (3.6,0.4)-- (3.6,0);
\draw (1.18,2.48)-- (0.9,2.04);
\draw (3,1.14)-- (2.72,0.7);
\draw (2.9,2.44)-- (3.16,2.12);
\draw (2.74,2.28)-- (3,1.96);
\draw (1.16,1.16)-- (1.42,0.84);
\draw (1,1)-- (1.26,0.68);
\begin{scriptsize}
\draw (-0.26,3.1) node {$A$};
\draw (-0.22,0) node {$D$};
\draw (4.25,0) node {$C$};
\draw (4.25,3.1) node {$B$};
\fill [color=uuuuuu] (2,1.5) circle (1.5pt);
\draw[color=uuuuuu] (2.06,1.9) node {$M$};
\end{scriptsize}
\end{tikzpicture}
\caption{An Example Geometric Figure \label{fig1}}
\end{center}
\end{figure}
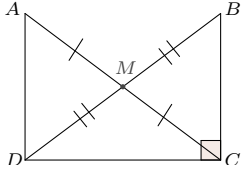

We illustrate the functionalities provided by GeoTutor through a concrete geometric problem taken from a textbook \cite{SinclairIX}.  Consider the geometric figure in Fig \ref{fig1} \cite{SinclairIX} and assume that we are given (i) $M$ is the midpoint of $\overline{AC}$, (ii) $M$ is the midpoint of $\overline{BD}$, and (iii) $m\angle BCD = 90^o$ \footnote{$m\angle BCD$ refers to the measure of $\angle BCD$}.

The following are some statements that are provable about Fig \ref{fig1} using these assumptions

\begin{multicols}{2}
\begin{enumerate}[A)]
\item $\Delta BMC \cong \Delta DMA$,
\item $m\angle ADC = 90^o$,
\item $\Delta ADC \cong \Delta BCD$,
\item $2\overline{BM}=\overline{AC}$, \footnote{$\overline{AC}$ is equal to twice $\overline{BM}$}
\item $\Delta DMC$ is isosceles,
\item $\Delta BMC$ is isosceles,
\item $\Delta DMA$ is isosceles,
\item $\overline{BC} \parallel \overline{AD}$. \footnote{$\overline{BC}$ is parallel to $\overline{AD}$}
\end{enumerate}
\end{multicols}

Only A) through D) are stated as problems (to prove) in \cite{SinclairIX}. When provided with this figure and assumptions (i) through (iii), GeoTutor will automatically synthesize proof problems that include all the  statements A) through H) above.

Beyond synthesizing problems based on the stated assumptions, GeoTutor will also synthesize all \emph{converse} proof problems. For example, in Fig \ref{fig1}, if  $\overline{BM} = \overline{DM}$ and $\Delta BMC$ is isosceles are given, one should be able to prove that $m\angle BCD = 90^o$; we call it a converse proof problem since the goal $m\angle BCD = 90^o$ is an assumption, specifically assumption (iii), in Fig \ref{fig1} while the fact $\Delta BMC$ is isosceles is goal F) in one of the original problems.

In addition, GeoTutor may classify two problems as \textit{analogous}; i.e., a student who is able to prove one should be able to prove the other. For example, in Fig \ref{fig1}, proving that F) $\Delta BMC$ is isosceles and G) $\Delta DMA$ are analogous in their statements and solutions. If a student struggles proving F), G) may be provided as a practice problem or as a means to master the proof technique.

GeoTutor can also classify synthesized problems as \emph{interesting} or \emph{uninteresting}. While statements such as $m\angle AMD = m \angle BMC$ can also be proven on Fig \ref{fig1}, GeoTutor would label such a problem as uninteresting since the given assumptions are not required to prove it. On the other hand, despite the fact that $2\overline{BM} = 4 \overline{AC}$ can be proved on Fig \ref{fig1} using the given assumptions, GeoTutor would label it as uninteresting since it can be derived from D) stated above through purely algebraic manipulation.

GeoTutor also supports \emph{queries} that a user can input to retrieve problems \emph{analogous} to a given problem, problems with a certain difficulty level (in terms of \emph{length} or \emph{width} of proofs or the axioms used), or problems based on the type of goal (proving congruence of triangles, equality of segments/angles, etc.).

If a student is stuck at a particular step in a proof, the system can provide a hint on the next step to follow (e.g., the correct axiom, definition, or theorem to be used). If a student is able to provide the complete proof for a problem, the system can grade it as correct or incorrect; in the latter case, it can point to the possible mistakes and suggest corrective steps or a more succinct solution.

The system provides an intuitive GUI interface leveraging LiveGeometry \cite{Live} and enables porting of the application to the web as well as deploy it as a standalone for use with a tablet. Currently, it synthesizes problems based on the immutable figure itself. For example, given an isosceles triangle, we may not alter the figure by making it scalene, nor changing any of the derivable characteristics such as the angle bisector of the angle opposite of the base intersecting the base at a point other than the midpoint. It is important to note that even though the user may provide an isosceles triangle, based on given assumptions, we may be able to prove that the triangle is, in fact, equilateral.

We evaluated GeoTutor on a corpus of high school geometry problems from standard geometry text books. Starting with a set of 155 textbook problems, GeoTutor automatically synthesized approximately 8000 related problems, and approximately 3000 converse problems, all of which were deemed interesting using our metric. Additionally, GeoTutor automatically classified the difficulty levels of these problems. Each problem instance required a few seconds to generate. Our evaluations indicate GeoTutor can be used as an effective component in computer-delivered personalized education in geometry education at the high school level.

\section{Theoretical Foundations for Problem Synthesis in Euclidean Geometry}\label{tf}
We currently consider immutable figures in which the properties of that figure are not allowed to be modified nor any new information constructed.
\subsection{Geometric Classes}
There are several distinct types of objects in Euclidean geometry, most notably: points, rays, segments, lines, triangles, quadrilaterals, and circles. For our purposes, we define a class for each geometric object: the class of points $\mathcal{P}$, the class of segments $\mathcal{S}$, the class of triangles $\mathcal{T}$, etc.

%

\subsection{Theories and Figures}
Let $\mathcal{L}$ be a logic \cite{CL97} in which properties of  a geometric figure are described.  We assume a finite set of geometric classes including point, segment, triangle, isosceles triangle, and equilateral triangle. Let $\mathscr{F} = \{ \mathcal{F}_1, \ldots, \mathcal{F}_k\}$ be the collection of $k$ geometric classes. Also let $F$ be a figure that belongs to a class $\mathcal{F}$: formally, $F \in \mathcal{F}$. We then define the theory of a class of figures  $\mathcal{F}$ as $Th(\mathcal{F}) = \{ \phi_1, \ldots, \phi_j \}$ where each $\phi_i$ is a property (an ${\cal L}$ formula) and $1 \leq i\leq j$, to be the minimal set of \textit{intrinsic} properties of $F$; i.e.,  $\forall \phi_i \in Th(\mathcal{F}), \{ Th(A_x) \cup Th(\mathcal{F}) \setminus \phi_i \} \nvDash \phi_i$ where $A_x$ is the set of Euclid's  axioms \cite{Jurgensen}.  That is, $Th(\mathcal{F})$ consists of all properties of a class $\mathcal{F}$ that are innate to the class, but cannot be proven. For example, for the  triangle class, one can neither prove that triangles have three segments nor prove that they have three internal angles. These are the intrinsic properties of the triangle class. 

\subsubsection{Ordering on Geometric Classes}

\begin{definition}  We define the ordering operator $\sqsubseteq$ on classes as $\mathcal{F}_1 \sqsubseteq \mathcal{F}_2$ if and only if $Th(\mathcal{F}_1) \vDash Th(\mathcal{F}_2)$, i.e., if $Th(\mathcal{F}_1)$ logically entails  $Th(\mathcal{F}_2)$.
\end{definition}
\begin{proposition} $\sqsubseteq$ defines a partial order on $\mathscr{F}$.
\end{proposition}
%
%

\begin{example} Considering the class of  triangles ($\mathcal{T}$), isosceles triangles ($\mathcal{I}$), and equilateral triangles ($\mathcal{E}$), it is clear $\mathcal{E} \sqsubseteq \mathcal{I} \sqsubseteq \mathcal{T}$ as $\mathcal{E}$ contains the most information and is thus the strongest class.
\end{example}
For a figure $F$ to be described by a particular class $\mathcal{F}$ we say that the figure forces the theory  of the class $\mathcal{F}$: $F \Vdash Th(\mathcal{F})$. Thus $F \in \mathcal{F}$ if and only if $F\Vdash Th(\mathcal{F})$. 

We now need to show that a figure cannot be an element in two distinct chains in the partial order; e.g. a figure cannot be both a triangle and circle.

\begin{lemma}Let $F$ be a figure and $\mathcal{F}_1, \mathcal{F}_2$ be classes. If $F \in \mathcal{F}_1$ and $F \in  \mathcal{F}_2$, then either $\mathcal{F}_1 \sqsubseteq \mathcal{F}_2$ or $\mathcal{F}_2 \sqsubseteq \mathcal{F}_1$.
\end{lemma} 

We also require a figure to be defined by the most appropriate class.

\begin{corollary} For a figure $F$, there exists classes $\mathcal{F}_b$ and $\mathcal{F}_B$ such that for $F \in \mathcal{F}_b, F \in \mathcal{F}_B$ and  for all $\mathcal{F}'$ such that $F \in \mathcal{F}'$, $\mathcal{F}_b \sqsubseteq \mathcal{F'} \sqsubseteq \mathcal{F}_B$.\end{corollary}

$\mathcal{F}_b$ defines the greatest lower bound of classes for a figure $F$. We call $\mathcal{F}_b$ the \textit{strongest} class corresponding to figure $F$ and write $\mathit{strong}(F)$.

$\mathcal{F}_B$ defines the least  upper bound of classes for a figure $F$. We call $\mathcal{F}_B$ the \textit{weakest} class corresponding to figure $F$ and write $\mathit{weak}(F)$.  

Given a geometric Figure $F$, we will construct two sets of properties that describe $F$. The first set of properties, $I$, describe the invariant characteristics of $F$. That is, we note the relationships among the points, lines, and shapes that are independent of specific information about $F$; that is, angles and distances between points may differ but not the overall structure of the figures.
Two figures $G$ and $G'$ are \textit{invariant} if there exists a class of geometric figures $\mathcal{C}$ such that $\mathit{weak}(G) = \mathcal{C} = \mathit{weak}(G')$. We write $G \approx_\mathcal{C} G'$ to say figure $G$ is invariant to figure $G'$ with respect to class $\mathcal{C}$. 

For a figure $F$, we define the theory of $F$ denoted by $Th(F)$ to be $Th(\mathcal{F})$ where $\mathcal{F}=\mathit{strong}(F)$.
 Let figure  $F$ be a right triangle $\Delta ABC$ with $m\angle BAC = 90^o$. The theory of $F$, denoted by $Th(F)$ is given by  
\begin{center}
\begin{align*}
Th(F) &= \{ \Delta ABC, m\angle BAC = 90^o \} \\
&= \{ \text{triangle}(A, B, C), \text{right\_angle}(B, A, C) \}
\end{align*}
\end{center}


Let $\mathcal{T}$ be the class of triangles and $\mathcal{T}_r$ be the class of right triangles, then we note for the right triangle  $F$  above,  it is true that  $F \in \mathcal{T}$ and $F \in \mathcal{T}_r$ with $\mathcal{T}_r \sqsubseteq \mathcal{T}$.
\subsubsection{Axioms}
For Euclidean geometry, we assume modified versions of Euclid's original axioms \cite{Jurgensen}; these axioms are universally quantified. Some of these axioms are stated below
\begin{enumerate}
\item Segment Addition: If $B$ is between $A$ and $C$, then $AB + BC = AC$.
\item  Algebraic Properties of equality including addition, subtraction, multiplication, and division.
\item  Equality ($=$). congruence ($\cong$), and similarity ($\sim$)  are equivalence relations.
\end{enumerate}
The set of axioms describing algebraic properties of equality including addition, subtraction, multiplication, and division and those describing the fact that equality ($=$). congruence ($\cong$), and similarity ($\sim$)  are equivalence relations are called the \emph{algebraic} axioms and are denoted by $A_a$. 
In addition, a few existentially quantified axioms are assumed. Examples are 
\begin{enumerate}
\item If two parallel lines are cut by a transversal, then corresponding angles are congruent.
\item SSS, SAS,  and ASA congruency of triangles.
\item Corresponding Parts of Congruent Triangles are Congruent (CPCTC).
\item AA Similarity of Triangles.
\end{enumerate}
Each of the axioms above requires an encoding into a logical form. Let's consider a few examples. Consider the Segment addition axiom. For this axiom to be applied, we require two distinct pieces of information: (1) three points are collinear, (2) which of the three points lies between the other two points.  Consider a segment $\overline{\upchi_1 \upchi_2}$ with point $\upchi_3$ between $\upchi_1$ and $\upchi_2$. Then

\begin{center}($\upchi_1, \upchi_2, \upchi_3$ collinear) $ \wedge$ ($\upchi_3$ between $\upchi_1$ and $\upchi_2) \Rightarrow \upchi_1\upchi_3 + \upchi_3\upchi_2 = \upchi_1\upchi_2$ \end{center}


With CPCTC, we require two congruent triangles.  We require the labeling of the respective vertices of the congruent triangles to be consistent.
\begin{align*}
(\Delta ABC) &\wedge (\Delta DEF) \wedge (\Delta ABC \cong \Delta DEF) \Rightarrow \\
& (\angle ABC \cong \angle DEF) \wedge (\angle BCA \cong \angle EFD) \wedge \\
& (\angle CAB \cong \angle FDE) \wedge (\overline{AB} \cong \overline{DE}) \wedge \\
& (\overline{BC} \cong \overline{EF}) \wedge (\overline{CA} \cong \overline{FD})\\
\end{align*}

\subsubsection{Definitions of Geometric Terms}
We presume standard definitions of common geometric terms; e.g.:

\begin{itemize}
\item \textit{Collinear} refers to a set of points lying on one line. 
\item \textit{Midpoint} of a segment refers to the point that divides a given segment into two congruent segments.
\end{itemize}

These definitions have ramifications because they imply more properties regarding a figure. For example, if $M$ is the midpoint of $\overline{XY}$, then the definition states $\overline{XM} = \overline{MY}$. However, the definitions are implicit in the theory of a figure $F$ as well as the theory of given information. For a figure $F$, we call this information the \emph{theory of assumptions}, $Th(A_s^F)$.
\subsection{Hypergraphs and Problems}
The formal framework that we use to represent a geometric figure together with the assumptions is a hypergraph. Proof problems will be synthsized by exploring this hypergraph. 
Given only a figure, we may construct a corresponding hypergraph based solely on the intrinsic properties, axioms, and student knowledge base. The student knowledge base comprises the lemmas and theorems that the student has proven so far. 

\begin{definition} [Basic Hypergraph] Given a figure $F$, the \textbf{\emph{basic hypergraph}} corresponding to $F$ is  $H_b^F(P, E)$ where $P$ is the set of nodes and $E$ is the set of hyperedges.  We define the set of  nodes in the hypergraph $P = Th(\mathcal{F}) \cup Th(A_x) \cup Th(K)$ where $\mathcal{F} = \mathit{strong}(F)$, $K$ is the student knowledge base, and $A_x$ is the set of Euclid's axioms. The hyperedges, $E$, of the the graph are defined as a set of functions mapping a set of  nodes to a single  node: $E \subseteq \bigcup_{i = 1}^{|P|} P^i \rightarrow P$ where $\langle p_1, \ldots, p_\ell \rangle \rightarrow p \in E$ if  $Th(\mathcal{F}) \cup Th(A_x) \models p_1 \wedge \ldots \wedge p_\ell \Rightarrow p$ holds true. \end{definition}

Each node in the basic hypergraph is \textit{typed} so it belongs to one discrete class in the set of types  $\tau = \{algebraic, geometric\}$. We make these distinctions among nodes so that later we may formally define a problem with respect to a basic hypergraph. We now define how the type of each node in a basic hypergraph is acquired.

\begin{definition}[Algebraic and Geometric Nodes] Let $n$ be a node in a basic hypergraph $H$. If $n$ is a propositional formula associated with some $a \in A_a$, we say $n$ is a \textbf{\emph{purely algebraic node}}. Let $H^T$ be the transpose of hypergraph $H$; that is, the nodes of $H^T$ are the same as those of  $H$ but with all edges in $H$ reversed. Define $\mathit{leaves}(H^T)$ to be the set of all nodes in $H^T$ without parents. If  for all $\ell \in \mathit{leaves}(H^T)$ such that there exists a path from $n$ to $l$ in $H^T$, $ell$ is a purely algebraic node,  we say $n$ is an \textbf{\emph{algebraic node}}. We note that purely algebraic nodes are considered algebraic nodes. We similarly define the terms \textbf{\emph{purely geometric nodes}} and \textbf{\emph{geometric nodes}} for Euclid's  axioms, $A_x$.\end{definition}
 

We can extend our notion of the basic hypergraph $H_b^F$ for a geometric figure $F$  by including the problem statement in the corresponding hypergraph. This is accomplished  by incorporating the assumptions, $A_s^F$, and the goal, $G$. 

\begin{definition}[Standard Hypergraph] Given a figure $F$ and corresponding basic hypergraph, $H_b^F(P,E)$, the \textbf{\emph{standard hypergraph}} corresponding to $F$ with assumptions $A_s^F$ and goal $G$, is given by $H_s^F(H_b^F, P_g, E_g, G)$. We define the additional set of typed nodes in the hypergraph $P_g = Th(A_s^F) \cup \{G\}$. The corresponding additional hyperedges, $E_g$ are a result of the theories derived from all typed nodes given by $P \cup P_g$ where $P$ are the typed nodes defined in $H_b^F$.\end{definition}

 If we do not distinguish between a basic hypergraph or standard hypergraph we will refer to a \textbf{\emph{problem hypergraph}}, $H(P, E)$ where $P$ is the set of typed nodes and $E$ is the set of hyperedges and when it is clear from cntext, we will simply call it a hypergraph.

It is clear that for a figure $F$, $H_b^F$ is a sub-hypergraph \cite{Berge} of  $H_s^F$.
\subsubsection{Geometry Problems}
A traditional high school geometry problem in simplest form is a natural language statement, but more common is the combination of a description composed of mathematical relationships and natural language which describe a figure. In a problem hypergraph, we informally define a problem as a set of typed nodes that describe the assumptions of the problem   and a corresponding typed goal node that follows from the assumptions. The corresponding path from the typed assumption nodes to the typed goal node is a solution to the problem (i.e., a proof of the goal).

\begin{definition} [Basic and Standard Problems] Given a basic hypergraph $H_b^F$ corresponding to a figure $F$, a \textbf{\emph{basic problem}} $P$ is a statement of the form $p_1 \wedge \ldots \wedge p_k \vdash p$ for some $k > 0$ where for all $i$, $p_i$ is the propositional formula corresponding to typed node $n_i$ of $H_b^F$, $p$ is the propositional formula corresponding to typed node $n$ of $H_b^F$, and there exists a path $\mathcal{P}$ from $\langle n_1, \ldots, n_k \rangle$ to $n$. The path $\mathcal{P}$ is a \textbf{\emph{solution}} to the problem. We say that $\mathscr{P}$ defines the collection of all paths in hypergraph $H_b^F$. A \textbf{\emph{standard problem}} is defined similarly for a standard hypergraph $H_s^F$. \end{definition}
We will use the general term \textit{problem} in situations where the context is clear.
For a goal $g$ and a set of source nodes $S$ in a standard hypergraph $H_s^F(H_b^F, P_g, E_g, G)$ corresponding to figure $F$ with assumptions $A$, we say that $S$ is \emph{minimal} with respect to $g$ if $S\vdash g$ is a problem and no $U \vdash g$ is a problem for $U \subset S$.

As mentioned in the Introduction, not all problems are interesting. Interesting problems for a figure and a set of assumptions are those that require at least one or more of the assumptions, the assumptions are minimal with respect to the goal, and the goal cannot be derived from a set of algebraic expressions through purely algebraic manipulation.

\begin{definition} [Interesting Problems] \label{definition} Let ${H_s}(H_b^F, P_g, E_g, G)$ be the standard hypergraph corresponding to figure $F$ with assumptions ${A_s}^F$. Let $P$ be a problem with source nodes $S$ used to construct $H_s^F$. Also let $g$ be the goal of $P$. We say that $P$ is an \textbf{\emph{interesting problem}} if (i) $S$ is a minimal set with respect to $g$, (ii) all direct predecessors of $g$ are not algebraic nodes, and (iii) $S \subseteq A_s^F$ where $|S \cap A_s^F| > 0$. That is, the problem must use at least one of the assumptions used to construct $H_s^F$. In case $S=A_s^F$, we call $P$ a \textbf{\emph{strictly interesting}} problem\end{definition}

\subsubsection{Analogous Problems}
We use the term \emph{analogous} to define a problem as an independent, 'interesting' problem that mimics the difficulty and length of a given problem.
For a problem $P$ in a hypergraph $H$, the problem hypergraph $\tilde{P}$ is the sub-hypergraph of $H$ induced by $P$.   We begin with a strict view of problem analogy that looks at problem hypergraphs as  graphs.
\begin{definition}[Coarse Problem Homomorphism] Let $H(V, E)$ and $H'(V', E')$ be problem hypergraphs. Then $\upphi: H \rightarrow H'$ is a \textbf{\emph{coarse problem homomorphism}} if for all $\langle v_1, \ldots, v_k \rangle = \vec{v} \in \mathcal{P}(V)$ and $v \in V$ such that $\vec{v} \rightarrow v \in E$,

\begin{enumerate}[(i)]
\item $v$ and $\upphi(v)$ are typed nodes in which $\mathit{type}(v) = \mathit{type}(\upphi(v)) \in \tau$,
\item $\vec{v}$ and $\upphi(\vec{v})$ are sets of typed nodes in which $|\vec{v}|_t = |\upphi(\vec{v})|_t$ for each type $t \in \tau$, and
\item there exists an edge $\upphi(\vec{v}) \rightarrow \upphi( \{ v \} ) \in E'$.
\end{enumerate}
\end{definition}
We then say $\upphi$ is a \textbf{\emph{coarse problem isomorphism}} if (i) $\upphi$ is a bijection, (ii) $\upphi$ is a coarse problem homomorphism, and (iii) $\upphi^{-1}$ is a coarse problem homomorphism. If $\upphi$ is a coarse problem isomorphism between $H$ and $H'$, we may write $H \cong_c H'$. We also say two problems $P_1$ and $P_2$ are \textbf{\emph{coarsely analogous}} if there exists a coarse problem isomorphism between $\tilde{P_1}$ and $\tilde{P_2}$. In Fig \ref{fig1}, the two problems proving that F) $\Delta BMC$ is isosceles and G) $\Delta DMA$ is isosceles are coarsely analogous. However, coarse analogy can be too strong a concept to formally capture the notion of ``analogy". To give an example, in Fig \ref{fig1}, a student who has been able to prove statements F) and G) should also be able to prove the statement E) since all three statements require one to prove that a particular triangle is isosceles  though the task of proving E)  is not coarsely analogous to that of proving F) or G).  Formally capturing a weaker notion of analogy motivates the following definition
\begin{definition}[Goal Analogous Problems] Let $P_1$ and $P_2$ be two problems with goals $g_1$ and $g_2$, respectively. Then we say problems $P_1$ and $P_2$ are \textbf{\emph{goal analogous problems}} if $type(g_1) = type(g_2)$. This is clearly an equivalence relation and we refer to the induced equivalence classes as \emph{goal analogous partition}.
\end{definition}

\section{Automatic Synthesis of Problems for Euclidean Geometry}\label{algorithm}
Given a problem hypergraph $H(P,E)$, Euclidean Geometry problems are automatically synthesized by using a pebbling technique \cite{DG84}.
\subsection{Pebbling Algorithm}
\begin{figure}
\begin{center}
\begin{tikzpicture}[line cap=round,line join=round,>=triangle 45,x=1.0cm,y=1.0cm, scale = 0.75]

\clip(-1.5,-1.5) rectangle (5.25,4.35);

\draw(0,3) circle (0.46cm);

\draw(2,3) circle (0.5cm);

\draw(4,3) circle (0.46cm);

\draw(0,-1) circle (0.46cm);

\draw(2,-1) circle (0.46cm);

\draw(4,-1) circle (0.46cm);

\draw(2,1) circle (0.5cm);

\draw [->] (2,2.5) -- (2.02,1.5);

\draw [->,dash pattern=on 3pt off 3pt] (2.02,-0.54) -- (2,0.5);

\draw (-1.25,4.50) node[anchor=north west] {\small SAS};

\draw [->,dash pattern=on 3pt off 3pt] (1.5,1.02) -- (0,2.54);

\draw [->,dash pattern=on 3pt off 3pt] (2.5,1) -- (4.02,2.54);

\draw (0, 4.2)-- (1.35, 4.2);

\draw (-1.25,4.01) node[anchor=north west] {\small ASA};

\draw [dash pattern=on 3pt off 3pt] (0,3.75)-- (1.4,3.75);

\draw (0.46,2.98)-- (2.01,2);

\draw (3.54,3)-- (2.01,2);

\draw [dash pattern=on 3pt off 3pt] (0.46,-1.02)-- (2.01,-0.02);

\draw [dash pattern=on 3pt off 3pt] (3.54,-1)-- (2.01,-0.02);

\draw [->] (2,0.5) -- (0,-0.54);

\draw [->] (2,0.5) -- (3.98,-0.54);

\draw [->,dash pattern=on 3pt off 3pt] (2.3,1.4) -- (2.3,2.6);

\draw [->] (2.3,0.6) -- (2.3,-0.65);

\draw (1.71,1.45) node[anchor=north west] {\scriptsize $\cong$};

\draw (1.60,1.1) node[anchor=north west] {\scriptsize $\Delta$'s};

\draw (4.15,1.25) node[anchor=north west] {\scriptsize $\Delta_2$};

\draw(4.5, 1) circle (0.5cm);

\draw (-1.1, 1.25) node[anchor=north west] {\scriptsize $\Delta_1$};

\draw(-0.75,1) circle (0.5cm);

\draw [dash pattern=on 3pt off 3pt] (4, 1) -- (2.01, 0);

\draw [dash pattern=on 3pt off 3pt] (-0.25, 1) -- (2.01,0);

\draw [-] (4, 1) -- (2.01, 2);

\draw [-] (-0.25,1) -- (2.01,2);

\end{tikzpicture}
\caption{SAS and ASA Forward- and Back-Hyperedges \label{fig7}} 
\end{center}

\end{figure}
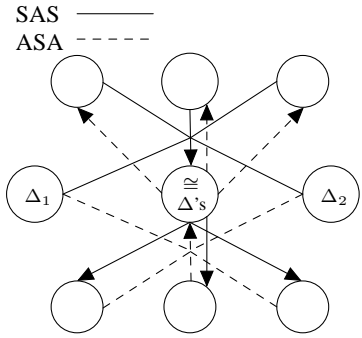

The representation of the knowledge base for each figure and set of assumptions is a hypergraph in which all hyperedges are many-to-one. There are two essential phases for the synthesis of geometry problems: construction and traversal of the hypergraph. Hypergraph construction is based on deductions resulting from standard Euclidean Geometry axioms, definitions, and theorems. Consider the case where we wish to prove that two triangles are congruent using the Side-Angle-Side (SAS) congruence axiom: two pairs of congruent sides and one pair of included congruent angles must relate the two respective triangles. A hyperedge may be constructed with the three congruence pairs as well as the two triangles as source nodes and the boolean fact that the two triangles are congruent as the target of the hyperedge as shown in Fig \ref{fig7}. Once the two triangles are determined to be congruent, the corresponding three parts of the congruent triangles (CPCTC) are then hyperedges with source being the congruent triangle node and target the CPCTC fact. It is important to note that for a complete hypergraph construction, other techniques to prove congruent triangles (Angle-Side-Angle, etc.) can then be applied to a subset of the six congruence pairs.

Given a complete hypergraph based on the figure and provided assumptions, the synthesis phase of the algorithm has several parts. Using a breadth-first pebbling, we identify forward edges and back-edges in the hypergraph. The forward pebbling mimics the construction of the hypergraph with the restriction that a hyperedge  from   a pebbled node to  another pebbled node implies a back-edge. Pebbling also excludes hyperedges based on any restrictions specified by the user query. That is, if a student wishes to practice problems related to congruent triangles, problems referencing similar triangles will not be generated. Any such restriction by the user query will therefore prohibit pebbling of hyperedges that are justified by axioms, definitions, and theorems related to similar triangles.

The second phase of pebbling seeks to identify all back-hyperedges. Based on the forward pebbled nodes, we pebble in the same breadth-first manner sorting the nodes from last deduced nodes to first. It is important to recognize that no deduced hyperedges will be both forward- and back-hyperedges except for hyperedges attributed to axioms. As an example of forward- and back-hyperedges, consider again the case of SAS congruence in Fig \ref{fig7}. The 5-to-1 hyperedge justified by SAS congruence is considered to be a forward edge. As the hypergraph is a complete representation of all facts from the figure and assumptions, there exists a hyperedge justified by Angle-Side-Angle (ASA) from the other three corresponding congruence pairs resulting from CPCTC. In this case, ASA is an example of a back-hyperedge in the graph.

\subsection{Problem Synthesis} \label{problem synthesis}

Given the set of  hyperedges of a hypergraph $H(P, E)$ acquired from pebbling, we can construct the associated set of all problems for a goal node. Consider a single hyperedge $e$ with source nodes $S$ and target $g$. For all $s \in S$, we recursively construct all problems $P_s$ with goal node $s$. Then, to acquire all problems $P_g$ with goal node $g$, we must compose $P_s$ for all $s \in S$ with $S \vdash g$. Formally,

\vspace{-0.5cm}

\begin{align*}
P_g =  \{ \left(U_1 \times \cdots \times U_{|S|} \right ) & \vdash g \mid \forall i, U_i \subseteq P \\
& \text{ and } U_i \vdash s \in P_s \text{ for some } s \in S \}.
\end{align*}

\section{Implementation and Evaluation} \label{implementation}
Since figures are immutable, we may leverage the coordinates of the input figure by calculating the ways in which the figure may be strengthened (a point in the middle of two points is a midpoint, a triangle is isosceles, etc). Each of the precomputed descriptors will be used as goals for forward problems as described in \ref{problem synthesis}. These pre-computed facts may not be found in the hypergraph since only deducible facts from the figure and assumptions are nodes in the hypergraph; for example, a midpoint may be implied by the coordinates of the figure, however, it is not a provable fact given the figure and assumptions. In the case of back-edges, we use the original problem assumptions as the respective goals. For a problem $P$ and its induced hypergraph $\tilde{P}$, a path from the source nodes to the goal node represents a solution to $P$. This solution can be matched to one developed by a student for grading purposes or can be used to provide a hint to a student stuck at a particular step.

\paragraph{Query-based Problem Synthesis}
GeoTutor supports the queries in Table \ref{query}; any query that requires stating the desired type of node is done so using an input menu by selecting, for example: parallel lines, congruent triangles, congruent angles, etc. For a problem $P$ with induced hypergraph $\tilde{P}$, we define proof width to be the width of $\tilde{P}$ \cite{MTT12} as well as define proof length to be the diameter of $\tilde{P}$. We define the number of deduced steps for a problem $P$ to be the number of hyperedges int $\tilde{P}$. 

\begin{table}

\begin{center}

\begin{caption}{Queries Supported by GeoTutor \label{query}}\end{caption}

\begin{tabular}{|c|c|}

\hline

Proof Width & $[a, b] : a, b \in \mathbb{Z}^+]$ \\

Proof Length & $[a, b] : a, b \in \mathbb{Z}^+]$ \\

Deductive Steps & $[a, b] : a, b \in \mathbb{Z}^+]$ \\

Source Type & Menu-Driven \\

Goal Type & Menu-Driven \\ \hline

\end{tabular}

\end{center}
\end{table}

\paragraph{Evaluation Methodology}\label{methodology}

For evaluation, we acquired 110 problems from standard mathematics textbooks in India for grades IX and X \cite{Sinclair,SinclairIX} as well as several recent textbooks and workbooks that are popular in the United States \cite{Boyd,Larson,Larsonpractice,Jurgensen}. Each textbook probem was hard-coded as a triple: $T = <F_T, A_T, G_T>$ where $F_T$ denotes the set of  intrinsic properties of the figure, $A_T$ denotes the assumptions as stated in the textbook, and $G_T$ the set of goals as stated in the textbook. For each textbook problem $T$, we generated the set of interesting problems for $T$, call it $\mathcal{P}_T$. We then validated for all $g \in G_T$ that there exists a problem $p \in \mathcal{P}_T$ such that for some $F \subseteq F_T$ where $F \neq \emptyset$,  $F \wedge A_T \vdash g$. The data based  on all such validated problems is summarized in  Tables \ref{general stats}, \ref{queries}, and \ref{interesting}.

\paragraph{Results}\label{results}

Table \ref{general stats} provides general cumulative statistics for all synthesized problems based on the 110 original textbook problems. For timing, all data were acquired on an Intel Core i5-2520M CPU at 2.5GHz with 8 GB RAM on 64-bit Windows 7 operating system and captured the entire process described in section \ref{algorithm} which includes coordinate-based computation, hypergraph construction, pebbling, and problem synthesis.

Table \ref{queries} demonstrates one method by which the set of interesting or strictly interesting problems can be partitioned: based on the number of deductive steps in a given problem. That is, the value 10.61 indicates, that, on average, a typical textbook problem results in approximately 10 interesting problems that require between 6 and 10 applications of axioms, theorems, or definitions to deduce the goal. This gives us a sense that, generally, all figures provide at least a medicum level of interesting problems at every level of difficulty.

Lastly, we note that there are many interesting problems that arise from a figure that are not \emph{strictly interesting} (see Definition  \ref{definition}). This type of problem is common in textbooks when a multi-stage problem occurs: part (a) may use one assumption and part (b) uses the remaining assumptions. This is the case in Fig \ref{fig1} where proving A) does not require assumption (iii). Table \ref{interesting} addresses this issue by partitioning the set of interesting problems using the percentage of givens as basis for comparison. Specifically, we look at 2.98 as the average number of problem which are interesting but use 25\% of the original textbook set of givens. We note that problems that require all the assumptions in the statement for their solution  are strictly interesting problems and that, on average, 19.7 strictly interesting problems are generated per textbook problem.

\begin{table}
\begin{center}
\begin{caption}{General Results Based on Textbook Problems \label{general stats}}\end{caption}
\begin{tabular}{|l|c|}
\hline
Figures & 110 \\ 
Original Textbook Problems & 155 \\ 
Generated Problems & 8307 \\ 
Interesting Problems & 4120 \\ 
Strictly Interesting Problems & 2168 \\ 
Converse Problems & 3097 \\ 
Time (secs / figure) & 3.90 \\ 
Ave. Goal Analogous Partitions & 9.04 \\
\end{tabular}
\begin{tabular}{|l|c|c|}
\hline
& Interesting & Strictly Interesting \\
Ave. Proof Width & 4.85 & 6.10 \\
Ave. Proof Length & 4.67 & 5.53 \\
Ave. Deductive Steps & 5.31 & 6.97 \\ \hline
\end{tabular}
\end{center}

\vspace{-0.65cm}

\end{table}


\begin{table}
\begin{center}
\begin{caption}{Difficulty-Based Partitioning Query \label{queries}}\end{caption}
\begin{tabular}{|c|c|c|}
\hline
Deductive & Ave. Interesting & Ave. Strictly \\
Steps & Problems & Interesting Problems \\
\hline
0-2 & 9.93 & 1.68\\
3-5 & 13.29 & 6.85 \\
6-10 & 10.61 & 7.84 \\
$>$ 10 & 3.63 & 3.34 \\
\hline
\end{tabular}
\end{center}
\vspace{-0.5cm}
\end{table}

\begin{table}
\begin{center}
\begin{caption}{Coverage of the Givens \label{interesting}}\end{caption}
\begin{tabular}{|c|c|c|c|c|}
\hline
0-25\% & 26-50\% & 51-75\% & 76-99\% & 100\% \\ \hline
2.98 & 11.13 & 3.63 & 0 & 19.7 \\ \hline
\end{tabular}
\end{center}
\vspace{-0.6cm}
\end{table}
\section{Related Work}
Existing automated tutoring systems provided by Wolfram Alpha \cite{wolfram}, Coursera \cite{Coursera}, and AutoTutor \cite{autotutor} cannot provide personalized feedback to the students. None of these systems can automatically synthesize analogous exercises to provide personalized practice to a student having difficulties in particular areas or types of exercises. Individualized, but analogous, assignments provided to students can mitigate cheating while maintaining fairness. None of the existent systems cover difficult topics like Geometry. Unlike GeoTutor, these systems provide problems from a predefined set that are slightly modified versions of those scoured from a plethora of textbooks. GeoTutor can synthesize problems beyond those available in textbooks; the student is free to generate their own problems by creating their own figures and associated assumptions. Altogether, it provides a personalized educational experience to the student that current systems do not.

Recently automatic problem generation has gained new interest with novel approaches in problem generation for natural deduction \cite{AGK13}, algebraic proof problems \cite{SGR12}, mathematical procedural problems \cite{AndersGP13}, embedded systems \cite{sadigh12}, Geometry constructions \cite{GKT11}, etc. All of them apply a similar technique: they first generalize an existing problem into a template, and then explore a space of possible solutions that fit this template. However, the specific approaches vary. Ahmed et al. build templates automatically, Andersen et al. and Singh et al. do it semi-automatically, and Sadigh et al. write templates manually. In contrast to this line of work, we do not use any manually written templates.

\section{Conclusions}\label{conclusions}
We described and evaluated  a technique for automatic problem synthesis for GeoTutor, an  intelligent tutoring system for Euclidean Geometry. GeoTutor has the ability to synthesize personalized assignments for students, generate analogous, but different,  exercises to curb cheating, and provide the educator with the ability to generate interesting assignments, exam questions, and their solutions automatically. Generating problems for assignments or exams is a difficult and tedious process for an educator and the gift of time for a teacher is the most valuable asset to educating all chidren. Time means more individual attention for each student so that teachers can do what they do best: teach students. In the future, we plan to deploy GeoTutor in high schools and conduct user studies to understand its effectiveness in an educational environment.

\newpage
\begin{small}
\bibliographystyle{aaai}
\bibliography{reference}
\end{small}

\end{document}